%% file: farinati.tex
\documentclass[runningheads]{llncs}
\usepackage[T1]{fontenc}
\usepackage{graphicx}
\usepackage{color}
\usepackage{hyperref}
\hypersetup{
    pdfborder={0 0 0}
}
\usepackage{amsmath}
\usepackage{cleveref}
\usepackage[numbers]{natbib}

\usepackage[linewidth=1pt,ticklabelscale=0.75,titleyshift=-1mm]{plotmacros}
\usetikzlibrary{external}

\definecolor{colDSurv}{RGB}{207,150,130}  
\definecolor{colRSF}{RGB}{127,201,172}      
\definecolor{colGA}{RGB}{244,177,131}       
\definecolor{colPSO}{RGB}{162,180,211}      
\definecolor{colOperon}{RGB}{178,210,143}   
\definecolor{colSLIM}{RGB}{250,210,107}     
\definecolor{colGPlearn}{RGB}{222,172,210}  

\definecolor{colCCI}{RGB}{200,50,50}        
\definecolor{colPCCI}{RGB}{70,130,200}      

\definecolor{colBinary}{RGB}{159,197,232}   
\definecolor{colContinuous}{RGB}{234,172,163} 

\definecolor{colSLIMperf}{RGB}{86,190,170}
\definecolor{colSLIMsmall}{RGB}{255,255,255}
\definecolor{colGPlearnPerf}{RGB}{244,164,120}
\definecolor{colGPlearnSmall}{RGB}{255,255,255}
\definecolor{colOperonPerf}{RGB}{140,170,220}
\definecolor{colOperonSmall}{RGB}{220,220,220}
\definecolor{colGApareto}{RGB}{240,160,220}
\definecolor{colPSOpareto}{RGB}{160,210,100}
\definecolor{colCCIpareto}{RGB}{180,60,60}
\definecolor{colPCCIpareto}{RGB}{70,130,230}
\definecolor{colRSFpareto}{RGB}{240,170,80}

\pgfplotsset{
    gridded/.style={
        grid=major,
        grid style={gray!30},
    },
    noinnerticks/.style={
        tick align=outside,
        tick pos=left,
    },
}

\pgfplotsset{
    baseBox/.style={
        boxplot,
        line width=1pt,
        boxplot/box extend=0.15,
    },
    binaryStyle/.style={
        baseBox,
        solid
    },
    continuousStyle/.style={
        baseBox,
        dash pattern=on 1.75pt off 1pt
    }
}

\usepackage[nohypertypes={acronym}]{glossaries}
\newacronym{pca}{PCa}{Prostate Cancer}
\newacronym[longplural={Comorbidities Indices}]{ci}{CI}{Comorbidities Index}
\newacronym{cci}{CCI}{Charlson Comorbidities Index}
\newacronym{rp}{RP}{Radical Prostatectomy}
\newacronym{coxph}{CPH}{Cox Proportional Hazards}
\newacronym{cindex}{CIndex}{Concordance Index}
\newacronym{pcci}{PCCI}{Prostate Cancer specific
Comorbidity Index}

\newacronym{ga}{GA}{Genetic Algorithm}
\newacronym{pso}{PSO}{Particle Swarm Optimization}
\newacronym{fstpso}{FST-PSO}{Fuzzy Self Tuning Particle Swarm Optimization}
\newacronym{gp}{GP}{Genetic Programming}
\newacronym{slim}{SLIM}{Semantic Learning through Inflate and deflate Mutation}
\newacronym{knn}{KNN}{K-Nearest Neighbors}
\newacronym{rsf}{RSF}{Random Survival Forest}
\newacronym{pbbia}{PBBIA}{Population Based Bio-Inspired Algorithm}
\newacronym{ea}{EA}{Evolutionary Algorithm}
\newacronym{so}{SO}{Swarm Optimization}
\newacronym{dsurv}{DeepSURV}{Deep Survival Unified Regression Variant}

\usepackage{caption}
\let\llncssubparagraph\subparagraph
\let\subparagraph\paragraph 
\usepackage[compact]{titlesec}         
\let\subparagraph\llncssubparagraph    
\titlespacing{\section}{0pt}{1pt}{1pt} 
\titlespacing{\subsection}{0pt}{1pt}{1pt}
\titlespacing{\subsubsection}{0pt}{1pt}{1pt}
\titlespacing{\paragraph}{0pt}{1pt}{1pt}

\usepackage{perpage}
\MakePerPage[1]{footnote}

\begin{document}
\title{Developing a novel Comorbidities Index for predicting 10-year mortality in Prostate Cancer patients: A computational data-driven approach}
\titlerunning{\textbf{Developing a novel 10-year CI for PCa patients}}

\author{Davide Farinati\inst{1} \and Francesco Barletta\inst{1} \and Paolo Zaurito\inst{1} \and Simone Scuderi\inst{1} \and Nicholas Raison\inst{2,3} \and Alejandro Granados\inst{2} \and Prokar Dasgupta\inst{2,3}\and Giorgio Gandaglia\inst{1} \and Alberto Briganti\inst{1}}
\authorrunning{D. Farinati et al.}
%
\institute{
Vita-Salute San Raffaele University, Milan, Italy
Comprehensive Cancer Center/Unit of Urology; URI; IRCCS Ospedale San Raffaele, Milan, Italy
 \\
 \email{\{farinati.davide, barletta.francesco, zaurito.paolo, scuderi.simone, gandaglia.giorgio, briganti.alberto\}@hsr.it}
 \and
 School of Biomedical Engineering \& Imaging Sciences, King’s College London, London, UK
 \\
 \email{\{nicholas.raison,alejandro.granados,prokar.dasgupta\}@kcl.ac.uk}
 \and
 Department of Urology, Guy’s and St Thomas’ NHS Foundation Trust, London, UK}

\maketitle              
\begin{abstract}

The \gls{cci} is a weighted additive index widely used to estimate ten-year mortality risk, but its original weights may not reflect contemporary prognoses. This limitation is critical in \gls{pca}, where radical treatment is recommended only for patients with a life expectancy of at least ten years. For candidates eligible for \gls{rp}, accurate estimation of ten-year other-cause mortality is essential to balance oncological benefit against competing risks and avoid overtreatment.
We propose a data-driven framework to derive a comorbidity index tailored to \gls{pca} patients considered for \gls{rp}. Using a retrospective single-institution cohort, we apply \glspl{pbbia} to recalibrate comorbidity weights and evolve alternative symbolic formulations optimized for ten-year survival discrimination. We compared six optimization strategies, including symbolic regression approaches based on \gls{gp}, population-based metaheuristics, clinically validated baselines, and survival prediction models.
Results show that \acrshort{ga}, \acrshort{fstpso}, and \acrshort{slim} outperform both the original \gls{cci} and the \gls{pcci}, particularly when \gls{pca}-specific variables are included, improving the \acrlong{cindex} by up to 0.1. GPLearn yields compact and interpretable models with competitive performance. Overall, the proposed approach provides an updated and interpretable tool to improve patient selection for \gls{rp}.

\end{abstract}

\glsresetall

\section{Introduction}
\label{sec: intro}

\gls{pca} is frequently characterized by a slow clinical course, and many patients with localized disease experience favorable long-term outcomes, with low ten-year cancer-specific mortality rates~\cite{Siegel2023CancerStats,Hamdy2016ProtecT}. 
Current clinical guidelines recommend radical treatment only for patients with an estimated life expectancy of at least ten years, underscoring the importance of accurately assessing competing mortality risks.

\gls{rp}, which consists of the complete surgical removal of the prostate gland along with selected surrounding tissues, is performed with curative intent in appropriately selected patients with localized \gls{pca}~\cite{EAU2024Guidelines}. 
As a major surgical procedure, \gls{rp} carries perioperative and long-term risks, making careful patient selection essential.

A key challenge in this context is the estimation of other-cause mortality in candidates eligible for \gls{rp}. 
While tumor characteristics are central to treatment decisions, comorbidity burden plays a critical role in determining whether a patient is likely to benefit from radical intervention. 
Despite guideline recommendations emphasizing life expectancy thresholds, there remains a lack of dedicated tools specifically designed to predict ten-year other-cause mortality in patients considered for \gls{rp} based on their comorbidity profile. 
To assist in this evaluation, clinicians use \glspl{ci}, which are quantitative instruments developed to estimate overall survival over a defined time horizon and to support individualized treatment decisions.

One of the most widely used \glspl{ci} is the \gls{cci}~\cite{cci}, originally developed to predict ten--year mortality risk based on a patient’s comorbid conditions. 
The \gls{cci} assigns a weighted score to each condition; for example, dementia is assigned one point, whereas leukemia is assigned two points. The total score is then used to estimate the patient’s ten--year survival probability\footnote{An online calculator implementing the \gls{cci} is available at \url{https://www.mdcalc.com/calc/3917/charlson-comorbidity-index-cci}.}.

Despite its widespread adoption, the \gls{cci} presents several limitations.
Developed in 1987, the index reflects the prognostic impact of comorbid conditions as understood at that time. 
As a result, certain diseases retain high weights that may no longer correspond to their current effect on mortality, given the substantial therapeutic advances achieved over the past decades. 
For example, HIV infection is assigned six points in the \gls{cci}. While this weighting was appropriate in the 1980s, when HIV was associated with very high mortality, the introduction and widespread use of combination antiretroviral therapy have transformed HIV into a chronic and manageable condition with significantly improved life expectancy~\cite{Palella1998HAART,Samji2013HIVLifeExpectancy}. 
Moreover, the \gls{cci} was originally designed as a general comorbidity index applicable across diverse patient populations, whereas different conditions are specific to certain patient populations.
In particular, \gls{pca} primarily affect older men, in whom the relative impact of specific comorbidities and competing mortality risks may differ substantially.

In this work, we propose a novel \gls{ci} specifically tailored to patients with \gls{pca}, with the aim of providing clinicians and surgeons a more reliable tool to guide the selection of candidates for \gls{rp}. To this end, we analyze a historical dataset comprising 2643 \gls{pca} patients treated between 1986 and 2025 at Ospedale San Raffaele (Milan, Italy), including detailed information on comorbidity conditions and ten--year survival outcomes from other cause mortality. 
We deliberately start by optimizing the CCI weights, as a weighted additive index remains directly interpretable by clinicians—each condition maps to an explicit point value—allowing us to improve predictive performance while retaining the transparency that motivates the index's clinical use.
In order to do s, we employ a range of computational methods to update the weights of the \gls{cci} and to investigate potential non--linear relationships between comorbidities and long--term survival.

First, to evolve the \gls{cci} weights, we evaluate a \gls{ga} and a \gls{fstpso} approach. Second, to investigate non-linear relationships we evaluate two \gls{gp} implementations, i.e., GPLearn and Operon, and one of its variants, i.e. \gls{slim}.
The main advantage of using \glspl{pbbia}~\cite{farinati_vanneschi_2024}
lies in their flexibility and ability to adapt naturally to different optimization tasks. 
In this work, we tailor these algorithms to the survival analysis setting by redefining their fitness function as the \gls{cindex}, a standard performance metric for censored time-to-event data. 
By directly optimizing the \gls{cindex}, both the weight vectors and the evolved symbolic expressions are guided toward maximizing discrimination of ten-year survival outcomes in \gls{pca} patients.

The paper is organized as follows, \Cref{sec: back} introduces the structure of the \gls{cci} and how it can be evaluated using the \gls{cindex}.
Subsequently \Cref{sec: lit rev} provides a brief review of existing \gls{ci} specific for \gls{pca} patients and applications of Machine Learning and Artificial Intelligence to the development of novel \gls{ci}.
Then \Cref{sec: exp meth} introduces the experimental methodology used in this study.
Finally \Cref{sec: exp res} presents the results obtained in this work.

\section{Background}
\label{sec: back}

In this Section we present some useful concepts that can help the reader fully grasp our study.

\subsection{Structure of the \gls{cci}}

The \gls{cci} assigns weighted scores to a predefined list of comorbidity conditions based on their association with ten-year mortality~\cite{cci}. 
The total score is obtained by summing the weights corresponding to the patient's conditions.
In this study we always employ an age-adjusted variant of the \gls{cci}, that was proposed by~\citet{charlson1994}.
This version of the \gls{cci} incorporates age as an additional scoring factor by adding one point for each decade of life beyond 50 years.
The weights assigned to each comorbidity can be found in \Cref{sec: exp res} in \Cref{tab:cci_variables_ga}.


The \gls{cci} score is calculated as:

\[
\text{\gls{cci}} = \sum_{i=1}^{k} w_i
\]

where $w_i$ represents the weight associated with each present comorbidity and $k$ is the total number of patient's condition.


In the original formulation, the \gls{cci} score was mapped to a predicted ten-year survival probability using an exponential model:

\[
\text{10-year survival probability} = 0.983^{\exp(0.9 \times \text{\gls{cci}})}
\]

This formulation reflects the exponential increase in mortality risk associated with higher comorbidity burden. The constant 0.983 represents the baseline ten-year survival probability in the reference population, and the factor $0.9$ scales the impact of each additional \gls{cci} point.


Contemporary implementations frequently replace the fixed exponential formula with a \gls{coxph} model~\cite{cox_oakes_2018}. 
In the \gls{coxph} framework, the \gls{cci} score (or its individual components) is included as a covariate in a semi-parametric regression model:

\[
h(t \mid X) = h_0(t)\exp(\beta \cdot \text{\gls{cci}})
\]

where $h(t \mid X)$ is the hazard at time $t$, $h_0(t)$ is the baseline hazard and $\beta$ is the regression coefficient estimated from the data.

This approach allows recalibration of weights to contemporary populations and avoids reliance on historical mortality rates from the 1980s. Moreover, individual comorbidities can be modeled separately, interactions can be explored, and nonlinear effects can be incorporated.

\subsection{\acrlong{cindex}}

The \gls{cindex} is a discrimination measure for censored survival data, and it commonly used to evaluate \glspl{ci}.
It measures the model’s ability to correctly rank patients according to risk.

Formally, the \gls{cindex} is defined as:

\[
C = \frac{\text{Number of concordant pairs}}{\text{Number of comparable pairs}}
\]

A pair of patients is concordant if the patient with the higher predicted risk experiences the event earlier than the patient with the lower predicted risk. Values range from 0.5, when the predictor is no better than a random prediction, 0.7--0.8, when the predictor has acceptable discrimination and  $>$0.8 when the predictor showcases strong discrimination.

In the context of comorbidity indices, the \gls{cindex} provides a quantitative measure of how well the index predicts mortality. When proposing a new comorbidity index or updating the \gls{cci} using \gls{coxph} or AI-based methods, improvement in \gls{cindex} is commonly used as a primary metric of enhanced predictive performance.

\section{Literature Review}
\label{sec: lit rev}

The \gls{cci} remains one of the most widely adopted tools for estimating long-term mortality risk in clinical populations. In \gls{pca}, accurate assessment of competing mortality risks is particularly important because the disease often follows an indolent course and many patients die from causes unrelated to cancer. 

\citet{pcci} proposed \gls{pcci}, an age-adjusted comorbidity index specifically designed to predict long-term other-cause mortality in men with localized \gls{pca}.
Their model incorporated both age and a broader set of comorbidities than those included in the \gls{cci}, adding conditions such as arrhythmia, valve disease, mild renal disease, other neurological disease, and inflammatory bowel disease, as well as finer severity gradations for conditions.
However, the weights were reweighted in a population
with prostate cancer to predict long-term mortality.

More recently, Sekine and Nakajima~\cite{acci} developed an algorithmic approach to compute the \gls{cci} using large-scale national registry data. By leveraging structured healthcare databases, they demonstrated how automated derivation of comorbidity scores can improve reproducibility and scalability. Algorithmic implementations of the \gls{cci} represent an important step toward integrating comorbidity assessment into real-world clinical decision support systems.

Beyond these approaches, contemporary research increasingly applies machine learning and artificial intelligence techniques to refine prognostic modeling. AI-based survival models can update comorbidity weights, model nonlinear interactions between diseases, and integrate high-dimensional clinical variables. Compared with the fixed additive structure of the original \gls{cci}, these models can potentially improve predictive accuracy while maintaining interpretability when properly validated. In \gls{pca}, such improvements are particularly relevant for estimating competing mortality and guiding treatment selection.

\section{Experimental Methodology}
\label{sec: exp meth}

In this Section we present the different datasets and algorithm used to model a novel \gls{ci}.
Additionally we also present for each algorithm the parameters employed.

\subsection{Experimental Feature Settings}
\label{sec: feat sec}

Three progressively enriched feature scenarios were defined to evaluate the impact of feature representation and model flexibility on predictive performance.

The first scenario replicates the standard age-adjusted \gls{cci}. It includes 23 binary variables: four age bands (50--59, 60--69, 70--79, 80+), weighted from 1 to 4 points, and 19 binary comorbidity components with their original published weights (1, 2, 3, or 6 points) as described by~\citet{cci} and presented in \Cref{tab:cci_variables_ga}.
Details about the patients cohort are presented in the Supplementary Material.

The second scenario extends the 23 CCI-age variables by incorporating approximately 69 additional pre-operative urological variables specific to \gls{pca} patients, yielding 92 features in total.
These include medications, lifestyle factors, laboratory values, hormonal markers, biopsy data, clinical staging variables, imaging findings, and functional scores. 
All additional variables were binarized using clinically meaningful thresholds (e.g., PSA $>$ 10 ng/mL, Gleason score $\geq 7$, BMI $\geq 25$ kg/m$^2$) or median-based splits when established cut-offs were unavailable. 
Variables already represented within the \gls{cci} components were excluded to prevent redundancy.
A table presenting all 69 \gls{pca} specific features is presented in the Supplementary Material.

The third scenario uses the same extended feature set but retains continuous variables in their original numerical form rather than binarizing them. 
This configuration enables algorithms capable of modeling nonlinear relationships to exploit the full information content of the data.
Algorithms capable of handling both binary and continuous variables, namely \gls{slim}, GPLearn and Operon, were trained and evaluated on both the extended binary and extended continuous representations. 


Binary variables were imputed using mode imputation (most frequent value), while continuous variables were imputed using \gls{knn} imputation with $k=5$. 
This approach preserves dataset dimensionality while minimizing bias introduced by incomplete observations.

\subsection{Optimization Methods}
\label{sec: optm meth}

We compared five optimization strategies, including symbolic regression approaches based on \gls{gp}, population-based metaheuristics, clinically validated baselines, and survival prediction models.

\subsubsection{Symbolic regression strategies}  Three \gls{gp}-based methods were employed: \gls{slim}, GPLearn, and Operon. 
For consistency, the function set and constants were identical across all \gls{gp} methods. 
The function set was $\{\text{add}, \text{subtract}, \text{multiply}, \text{protected division}\}$, and the available constants were $\{2, 3, 4, 5, -1\}$. 
All symbolic models were constrained to a maximum expression length of 30.
The parameters of the \gls{gp} based methods are presented in \Cref{tab: gp param}.

\begin{table}[h]
\centering
\begin{tabular}{lcc}
\hline
\textbf{Parameter} & \textbf{GP Methods} & \textbf{\gls{slim}} \\
\hline
Population size & 200 & 200 \\
Generations & 100 & 100 \\
Max expression length & 30 & None \\
Max Initial depth & 6 & 6 \\
Max depth & 15 & None \\
Parsimony coefficient (GPlearn) & 0.001  & -- \\
$p_{\text{inflate}}$ & -- & 0.3 \\
$p_{\text{deflate}}$ & -- & 0.7 \\
\hline
\end{tabular}
\caption{Hyperparameters for symbolic regression methods.}
\label{tab: gp param}
\end{table}

\gls{slim}, a promising variant of \gls{gp} employing Geometric Operators~\cite{farinati_pietropolli_vanneschi_2025}, was implemented using the \texttt{slim\_gsgp}~\cite{slim, slim_impl}\footnote{https://github.com/DALabNOVA/slim/tree/main/slim\_gsgp} library, specifically the variant +SIG1.
GPLearn~\cite{lopez_2026}\footnote{https://gplearn.readthedocs.io/en/stable/} was implemented through the \texttt{SymbolicRegressor} interface.
Both \gls{slim} and GPLearn were evolved with a custom \gls{cindex} fitness function.
Operon~\cite{operon}\footnote{https://github.com/heal-research/operon} was implemented using \texttt{operon.sklearn}.
Since custom survival losses are not supported, it was trained on survival duration as the optimization target.
All three algorithms used the default settings that can be found in the respecitive libraries.

\subsubsection{Population-based metaheuristics} 
The choice of GA and FST-PSO is motivated by three considerations. First, the optimization target—the \gls{cindex}—is a rank-based, non-differentiable metric that rules out gradient-based fitting and favors population-based search, which also keeps these methods directly comparable to the GP-based approaches under a common search paradigm. 
Second, the two algorithms recover complementary forms of the index: the GA operates on integer weights, reproducing the discrete point-score structure clinicians already use with the CCI, while FST-PSO explores continuous weights and removes the need for manual tuning through its fuzzy self-adaptive control.
Third, both preserve the additive, weight-based formulation: rather than immediately adopting interaction-aware models at the cost of transparency, we first assess whether recalibrating the existing weights to a contemporary cohort improves discrimination, keeping each comorbidity mapped to an explicit contribution and establishing an interpretable baseline for the symbolic-regression models that follow.

A custom \gls{ga} was implemented to optimize integer weight vectors for a weighted linear comorbidity score. Each individual in the population is represented as a vector of integers (one weight per variable, range 0--9). The phenotype corresponds to a linear risk score obtained as the weighted sum of the selected features, and fitness is defined as the \gls{cindex} computed between the resulting risk score and observed survival data.

Parent selection is performed using tournament selection with pool size two.
Recombination is implemented via uniform crossover with $p_c = 0.6$, where a randomly generated binary mask determines, gene by gene, which parent contributes each position in the offspring. 
Mutation is performed through one-unit perturbation $p_m = 0.1$, where each gene has a 10\% probability of being incremented or decremented by one unit.
This small-step mutation encourages local refinement of weights and stabilizes convergence by preventing disruptive jumps in the search space. 
Elitism is enabled, meaning that the best individual in each generation is copied unchanged into the next generation, guaranteeing monotonic non-decreasing best fitness across generations.

When applied to the standard 23-variable \gls{cci} feature set, the original Charlson weight vector~\cite{cci} is injected as one individual in the initial population.
This seeding strategy provides a clinically meaningful starting point and allows the evolutionary search to explore refinements around an already validated solution rather than starting purely at random.

\gls{pso} was implemented using \gls{fstpso}~\cite{fst_pso}\footnote{https://github.com/aresio/fst-pso} to optimize continuous weight vectors within the range $[0,10]$ per variable, and was applied to both the standard and extended binary feature sets. 
Unlike classical \gls{pso}, which requires manual tuning of the inertia weight and acceleration coefficients, \gls{fstpso} employs a fuzzy logic controller to adjust these parameters dynamically based on swarm behavior, improving robustness across different feature dimensionalities without manual calibration.

The parameters of \gls{ga} and \gls{fstpso} are presented in \Cref{tab: gapso param}.

\begin{table}[h]
\centering
\begin{tabular}{lcc}
\hline
\textbf{Parameter} & \textbf{\gls{ga}} & \textbf{\gls{fstpso}} \\
\hline
Population size & 200 & -- \\
Generations / Iterations & 100 & 100 \\
Weight range & [0,$+\infty$) & [0,10] \\
Crossover probability & 0.6 & -- \\
Mutation probability & 0.1 & -- \\
\hline
\end{tabular}
\caption{Hyperparameters for population-based metaheuristics.}
\label{tab: gapso param}
\end{table}

\subsubsection{Clinically validated baselines} Two clinically grounded baselines were included for comparison: the original age-adjusted \gls{cci}~\cite{cci} and the \gls{pcci} proposed by \citet{pcci}, which provides validated weights for predicting other-cause mortality in \gls{pca} patients.

\subsubsection{Survival prediction models} In order to evaluate the performance of the used models, we employed two \gls{ci} baselines, namely \gls{cci} and \gls{pcci}, as well as \gls{rsf}~\cite{rsf} and \gls{dsurv}~\cite{deepsurv}. \gls{rsf} is a black-box extension of random forests tailored for right-censored time-to-event data, in which an ensemble of survival trees is grown on bootstrap samples using survival-specific splitting rules, and predictions are obtained by aggregating survival estimates across the forest. \gls{dsurv} is a deep learning-based model that captures complex nonlinear relationships between covariates and survival outcomes using a neural network trained with a Cox partial likelihood loss.



\section{Experimental Results}
\label{sec: exp res}

This Section presents the experimental results obtained by this work.
At first, in \Cref{sec: res std} we evaluate the recalibration of the original \gls{cci} weights, exploring also non linear combinations.
Then, in \Cref{sec: res pca} we focus on the results obtained when extending the input dataset with a set of new variables specific to \gls{pca} patients.
Finally in \Cref{sec: res int} we focus on interpreting the results obtained by the different methods when applied to the extended variable set.

Model evaluation was conducted using Monte Carlo cross-validation (also known as repeated random sub-sampling validation) with 30 independent runs. 
In each run, the dataset was randomly partitioned into an 80\% training and a 20\% test subset, with the partition regenerated independently for every run and the proportion of survival events preserved across splits.
Unlike a single hold-out, this repeated-resampling scheme yields a distribution of test-set scores rather than a single estimate, which is why all results are reported as distributions across the 30 runs.
For each analysis the p-values resulting from a Wilcoxon rank-sum test can be found in the Supplementary Material.

\subsection{Recalibration of the Original \gls{cci} Weights}
\label{sec: res std}

\Cref{fig:boxplot_all_standard} shows the distribution of the test-set \gls{cindex} across all compared methods.

\input{figures/boxplot_all_standard}

We observe that \gls{rsf}, \gls{ga}, and \gls{slim} are the only approaches that consistently outperform both the original \gls{cci} and the prostate-adapted \gls{pcci}. 
Among these, \gls{rsf} achieves the highest overall performance, while \gls{slim} and \gls{ga} exhibit comparable results. 
In contrast, \gls{fstpso}, \gls{dsurv} and GPLearn perform similarly to each other and remain close to the baseline indices (\gls{cci} and \gls{pcci}).
Operon shows the weakest performance, likely reflecting the limitation of optimizing survival time directly rather than using the \gls{cindex} as the fitness function.

\Cref{fig:pareto_standard} illustrates the trade-off between predictive performance (test \gls{cindex}) and model complexity (solution size) in a two-dimensional space.

\input{figures/pareto_standard}

For \gls{slim}, GPLearn, and Operon, both the best-performing solution and the most compact solution are displayed. 
For the remaining methods, only the best-performing solution is shown.
For the indices (\gls{cci} and \gls{pcci}) and for the weight-optimization methods (\gls{ga} and \gls{fstpso}), model size is computed from the corresponding linear expression combining all of the optimized weights, including the ones to which a 0 was aasigned.
Since \gls{rsf} and \gls{dsurv} are a black-box models, they are represented as a horizontal line corresponding to their best observed performance.

From this plot, we observe that \gls{slim} and \gls{rsf} achieve similar peak predictive performance.
However, GPLearn is capable of producing substantially more compact symbolic solutions while sacrificing only a small amount of predictive accuracy, having similar performance to \gls{dsurv}.
This highlights a favorable interpretability--performance trade-off.
Finally also \gls{ga} reaches good performance, lying on the Pareto front together with \gls{slim} and GPLearn.

Table~\ref{tab:cci_variables_ga} reports the median weights, with the corresponding IQR, of the final evolved individuals across all \gls{ga} runs, compared to the original age-adjusted \gls{cci} weights.

\begin{table}[h]
\centering
\begin{tabular}{lcc}
\hline
\textbf{Comorbidity / Age Band} & \textbf{Original Weight} & \textbf{GA Median (IQR)} \\
\hline
\multicolumn{3}{l}{\textit{Age Bands (Age-Adjusted CCI)}} \\
50--59 years & 1 & 0 (0--0) \\
60--69 years & 2 & 0 (0--0) \\
70--79 years & 3 & 5.5 (3.0--7.0) \\
80+ years & 4 & 11.0 (10.0--12.0) \\
\hline
\multicolumn{3}{l}{\textit{Comorbidities}} \\
Myocardial infarction & 1 & 6.0 (4.2--7.0) \\
Congestive heart failure & 1 & 0 (0--2.2) \\
Peripheral vascular disease & 1 & 0 (0--0) \\
Cerebrovascular disease & 1 & 0 (0--0) \\
Dementia & 1 & 10.0 (9.0--11.0) \\
Chronic pulmonary disease & 1 & 2.0 (0.2--4.0) \\
Connective tissue disease & 1 & 0 (0--0) \\
Peptic ulcer disease & 1 & 0 (0--2.0) \\
Mild liver disease & 1 & 2.0 (0.2--2.0) \\
Diabetes without end-organ damage & 1 & 2.5 (2.0--4.0) \\
Hemiplegia & 2 & 4.0 (1.2--5.8) \\
Moderate or severe renal disease & 2 & 0 (0--6.0) \\
Diabetes with end-organ damage & 2 & 5.0 (2.0--7.0) \\
Any tumor (within 5 years) & 2 & 2.0 (2.0--4.0) \\
Leukemia & 2 & 0 (0--0) \\
Lymphoma & 2 & 0.5 (0--6.0) \\
Moderate or severe liver disease & 3 & 8.0 (7.0--9.0) \\
Metastatic solid tumor & 6 & 5.0 (1.0--8.0) \\
AIDS/HIV & 6 & 0.5 (0--2.8) \\
\hline
\end{tabular}
\caption{Age bands and comorbidities with original age-adjusted \gls{cci} weights and the optimized \gls{ga} weights, reported as the median (IQR) across runs.}
\label{tab:cci_variables_ga}
\end{table}

Notably, the optimized weights deviate substantially from the original clinical scoring scheme. 
The \gls{ga} assigns markedly higher importance to advanced age (median 11.0 for the 80+ band and 5.5 for 70--79) and to conditions such as dementia, severe liver disease, and myocardial infarction, while driving several variables, including peripheral vascular disease, leukemia, and the younger age bands, toward zero. 
These weights must be read in light of the cohort's structure, presented in the Supplementary Material: only 0.3\% of patients fall in the 80+ band, and most comorbidities are present in fewer than 2\% of patients, with some, such as leukemia (0.2\%) and congestive heart failure (0.1\%), observed in only a handful of cases. 
Many weights are therefore estimated from very few events and are highly unstable, as reflected in the wide interquartile ranges. 
Apparent inversions, where a rare severe comorbidity receives a near-zero median weight while an age band receives a large one, are most plausibly an artifact of this low support in a relatively healthy cohort rather than a genuine reversal of clinical risk.

\subsection{Integration of \gls{pca}-Specific Clinical Variables}
\label{sec: res pca}

\Cref{fig:boxplot_urological} presents the test \gls{cindex} of all methods when including \gls{pca} specific variables in the dataset.
For the models that can be adapted to either the continuous set or the binary set, the best version according to the performance on the train set, is included, the in depth comparison can be found in the Supplementary Material.

\input{figures/boxplot_urological}

We can observe how \gls{ga}, \gls{fstpso} and \gls{slim} are the only methods that outperform \gls{cci} and \gls{pcci}, being \gls{ga} the best performing one.
\gls{rsf}, \gls{dsurv} and GPLearn have comparable performance between themselves and \gls{cci} and \gls{pcci}, meanwhile Operon and \gls{dsurv} are the worst performing models.

Since for this work we also highly value interpretability, we display the model size in \Cref{fig:boxplot_size_urological}.

\input{figures/boxplot_size_urological}

From this Figure we can observe how GPLearn and Operon are able to produce the smallest solutions, even smaller that the linear models used by \gls{cci} and \gls{pcci}.

Finally we display the trade-off between predictive performance (test \gls{cindex}) and model complexity (solution size) in a two-dimensional space in \Cref{fig:pareto_urological}, following the same criteria as \Cref{fig:pareto_standard}.

\input{figures/pareto_urological}

From this plot we can observe how \gls{fstpso}, and both the solutions of \gls{slim} and GPLearn lie on the Pareto front.
\gls{fstpso} producing the biggest solutions and GPLearn the smallest ones.
If a single model were to be selected with both criteria in mind, the most suitable candidate would likely be the best-performing solution generated by GPLearn. 
This model achieves competitive predictive performance—outperforming most approaches, with the exception of \gls{slim}, \gls{ga}, and \gls{fstpso}—while remaining highly compact, with slightly fewer than 50 nodes. 
This balance between discrimination ability and structural simplicity makes it an attractive compromise between accuracy and interpretability.

\subsection{Model Interpretability and Clinical Insight}
\label{sec: res int}

In the final section of the Experimental Results we focus on the clinical interpretability of the models produced in this work.
Specifically we will display the smallest final solution returned by \gls{slim} and the best performing one returned by GPLearn.
These two have a test \gls{cindex} of 0.657 and 0.63 respectively, outperforming all other methods except \gls{ga} and \gls{fstpso}.

The symbolic expression returned by \gls{slim} is composed of 101 nodes and can be represented as:

\begin{equation}
\resizebox{0.9\textwidth}{!}{$
\begin{aligned}
F =
&\, \mathrm{NeoadjuvantHormonalTherapy}
+ \mathrm{Pacemaker}
\\[6pt]
&+ 0.9541231144 \cdot
\tanh\!\left(\frac{\mathrm{DiureticTherapy} + \mathrm{SolventExposure}}{2}\right)
\\
&+ 0.5116465795 \cdot
\tanh\!\left(\frac{\mathrm{HighRiskGroup} - \mathrm{Leukemia}}{2}\right)
\\
&+ 0.4138471465 \cdot
\tanh\!\left(\frac{\mathrm{Age70to79} - \mathrm{PulmonaryEmbolism}}{2}\right)
\\
&+ 0.5865985481 \cdot
\tanh\!\left(\frac{\mathrm{AlcoholAbuseIndex} - 2}{2}\right)
\\
&+ 0.5465666532 \cdot
\tanh\!\left(\frac{\mathrm{Age70to79} - \mathrm{CardiacArrhythmia}}{2}\right)
\\
&+ 0.7559014451 \cdot
\tanh\!\left(\frac{1 + \mathrm{Depression}}{2}\right)
\\
&+ 0.5669939785 \cdot
\tanh\!\left(\frac{\mathrm{SmokingBurden}}{2 \cdot \mathrm{Age50to59}}\right)
\\
&+ 0.6346833204 \cdot
\tanh\!\left(\frac{\mathrm{Hypertriglyceridemia}}{2 \cdot \mathrm{PreviousRectalTumor}}\right)
\\
&+ 0.9540509233 \cdot
\tanh\!\left(\frac{\mathrm{OralAnticoagulantTherapy} + \mathrm{PreoperativePET}}{2}\right)
\end{aligned}$}
\label{eq:slim_rewritten}
\end{equation}

Meanwhile, the one produced by GPLearn is composed of 31 nodes and can be represented as:

\begin{equation}
\resizebox{0.9\textwidth}{!}{$
\begin{aligned}
F =\;&
\frac{
\mathrm{AntihypertensiveTherapy}
+ \mathrm{CardiacArrhythmia}
+ \mathrm{DiureticTherapy}
+ \mathrm{PsychotropicMedication}
}{
\frac{\mathrm{SecondaryGleasonPattern}}{\mathrm{SmokingBurden}}
+
\frac{\mathrm{PreviousNephrectomy}}{\mathrm{SolventExposure}}
}
\\[6pt]
&-
\Big(
\mathrm{PeripheralVascularDisease}
+ \mathrm{SmokingStatus} \\
& + \mathrm{AntianginalTherapy}
+ \mathrm{HemorrhoidalDisease}
-
\mathrm{PepticUlcerDisease}
\\
&\quad
- \mathrm{NeoadjuvantHormonalTherapy}
- \frac{\mathrm{AntiplateletTherapy}}{\mathrm{RiskGroup}}
\Big)
\end{aligned}
$}
\label{eq:gplearn_rewritten}
\end{equation}
Given the usage of sigmoid in the formula of the Geometric Operators employed by \gls{slim}~\cite{slim}, its solutions lack interpretability.
Meanwhile, the solution produced by GPLearn has potential to be interpreted by clinicians.


\section{Conclusion}
\label{sec: conc}

This work introduces a data-driven framework for developing a novel \gls{ci} tailored to patients with \gls{pca}. 
Given that current guidelines recommend radical treatment only for individuals with a life expectancy of at least ten years, accurately predicting ten-year other-cause survival is crucial when evaluating candidates for \gls{rp}. 
By focusing specifically on the estimation of ten-year mortality from competing health conditions, the proposed model aims to support clinicians in identifying patients who are most likely to benefit from surgery, while avoiding overtreatment in those with limited life expectancy.

We evaluated several \glspl{pbbia}, comparing their performance with established \glspl{ci}, namely \gls{cci} and \gls{pcci}, as well as with survival-specific machine learning methods, \gls{rsf} and \gls{dsurv}. 
Two complementary strategies were considered, at first optimizing the weights of the original \gls{cci} framework, and then evolving novel symbolic expressions to derive alternative formulations of the \gls{ci}. 
Furthermore, we investigated the impact of incorporating \gls{pca}-specific clinical variables into the feature set.

The results demonstrate that \acrshort{ga}, \acrshort{fstpso}, and \acrshort{slim} are capable of generating optimized weight configurations and symbolic models that outperform existing \glspl{ci} and, in some cases, established survival analysis approaches. 
In addition, GPLearn proved particularly effective at producing compact and interpretable symbolic expressions while maintaining competitive predictive performance.

In this work, \glspl{pbbia} offered an interpretable and flexible framework for the development of \glspl{ci} tailored to patients with \gls{pca}. 
Future research should aim to validate and extend the proposed methodology on external cohorts from other institutions to assess its robustness across different patient populations.
Moreover, applying this framework to other clinical conditions would allow further evaluation of its generalization capability and its potential as a broadly applicable tool for comorbidity-based risk stratification.

\paragraph{Supplementary Material} The Supplementary material of this study is available at \url{https://github.com/Fari98/PPSN26_Developing_a_Novel_CIndex_Supplementary_Material}.

\newpage

\bibliographystyle{plainnat}
\bibliography{bibl}

\end{document}

%% file: figures/boxplot_all_standard.tex
\begin{figure*}[!h]
    \centering
    \begin{tikzpicture}
        \begin{axis}[
            width=\textwidth,
            height=0.5\textwidth,
            gridded,
            noinnerticks,
            ylabel={Test C-index},
            ymin=0.37, ymax=0.67,
            xtick=\empty,
            xmin=-.7, xmax=7.7,
            boxplot/draw direction=y,
            boxplot/every outlier/.style={mark=*, mark size=1.5pt, mark options={fill=gray!70, draw=gray!70}},
        ]

          \addplot[boxplot prepared={
              lower whisker=0.52748,
              lower quartile=0.56336,
              median=0.58360,
              upper quartile=0.59269,
              upper whisker=0.62162,
          }, ,fill=colDSurv!25, draw=colDSurv, boxplot/draw position=0] coordinates {};

        \addplot[boxplot prepared={
            lower whisker=0.575,
            lower quartile=0.595,
            median=0.607,
            upper quartile=0.615,
            upper whisker=0.635,
        }, fill=colRSF!25, draw=colRSF, boxplot/draw position=1] coordinates {(1,0.65)};

        \addplot[boxplot prepared={
            lower whisker=0.575,
            lower quartile=0.595,
            median=0.601,
            upper quartile=0.608,
            upper whisker=0.62,
        }, fill=colGA!25, draw=colGA, boxplot/draw position=2] coordinates {(2,0.635)(2,0.637)};

        \addplot[boxplot prepared={
            lower whisker=0.51,
            lower quartile=0.535,
            median=0.555,
            upper quartile=0.565,
            upper whisker=0.59,
        }, fill=colPSO!25, draw=colPSO, boxplot/draw position=3] coordinates {(3,0.605)(3,0.63)(3,0.484)};

        \addplot[boxplot prepared={
            lower whisker=0.49,
            lower quartile=0.535,
            median=0.555,
            upper quartile=0.58,
            upper whisker=0.615,
        }, fill=colGPlearn!25, draw=colGPlearn, boxplot/draw position=4.5] coordinates {(4.5,0.475)};

        \addplot[boxplot prepared={
            lower whisker=0.38,
            lower quartile=0.40,
            median=0.425,
            upper quartile=0.445,
            upper whisker=0.485,
        }, fill=colOperon!25, draw=colOperon, boxplot/draw position=5.5] coordinates {};

        \addplot[boxplot prepared={
            lower whisker=0.575,
            lower quartile=0.595,
            median=0.605,
            upper quartile=0.615,
            upper whisker=0.635,
        }, fill=colSLIM!25, draw=colSLIM, boxplot/draw position=6.5] coordinates {};

        \addplot[colCCI, densely dashed, thick] coordinates {(-3,0.549) (7.7,0.549)};
        \addplot[colPCCI, densely dashed, thick] coordinates {(-3,0.560) (7.7,0.560)};

        \end{axis}
    \end{tikzpicture}

    \addlegendimageintext{draw=colDSurv, fill=colDSurv!25, area legend, line width=1pt} DeepSurv
    \addlegendimageintext{draw=colRSF, fill=colRSF!25, area legend, line width=1pt} RSF
    \addlegendimageintext{draw=colGA, fill=colGA!25, area legend, line width=1pt} GA
    \addlegendimageintext{draw=colPSO, fill=colPSO!25, area legend, line width=1pt} FST-PSO 
    \addlegendimageintext{draw=colGPlearn, fill=colGPlearn!25, area legend, line width=1pt} GPlearn \\
    \addlegendimageintext{draw=colOperon, fill=colOperon!25, area legend, line width=1pt} Operon
    \addlegendimageintext{draw=colSLIM, fill=colSLIM!25, area legend, line width=1pt} SLIM
    \hspace{3mm} 
    \addlegendimageintext{draw=colCCI, densely dashed, line width=1.5pt} Standard CCI
    \addlegendimageintext{draw=colPCCI, densely dashed, line width=1.5pt} PCCI

    \caption{Test C-index comparison across all methods on the standard dataset.}
    \label{fig:boxplot_all_standard}
\end{figure*}
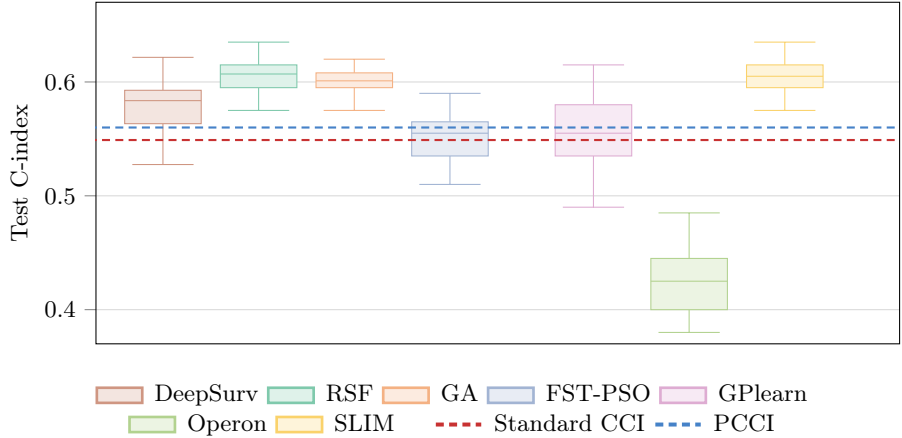

%% file: figures/pareto_standard.tex
\begin{figure*}[!h]
    \centering
    \begin{tikzpicture}
        \begin{axis}[
            width=\textwidth,
            height=0.5\textwidth,
            gridded,
            noinnerticks,
            xlabel={Solution Size (nodes)},
            ylabel={Test C-index},
            ymin=0.44, ymax=0.66,
            xmin=-5, xmax=110,
            x dir=reverse,
            legend style={hide legend},
        ]

        \addplot[colSLIM!50, dash pattern=on 4pt off 2pt, semithick, forget plot]
            coordinates {(100,0.649) (91.5,0.616)};
        \addplot[colGPlearn!50, dash pattern=on 4pt off 2pt, semithick, forget plot]
            coordinates {(60,0.617) (10,0.545)};
        \addplot[colOperon!50, dash pattern=on 4pt off 2pt, semithick, forget plot]
            coordinates {(58,0.497) (44,0.457)};

        \addplot[colRSF, dash pattern=on 6pt off 4pt, very thick]
            coordinates {(-5,0.649) (110,0.649)};
        \addplot[colDSurv, dash pattern=on 6pt off 4pt, very thick]
            coordinates {(-5,0.621) (110,0.621)};

        \addplot[only marks, mark=*, mark size=5pt, fill=colSLIM, draw=colSLIM]
            coordinates {(100,0.649)};
        \addplot[only marks, mark=square*, mark size=4.5pt, fill=colGPlearn, draw=colGPlearn]
            coordinates {(60,0.617)};
        \addplot[only marks, mark=diamond*, mark size=5.5pt, fill=colOperon, draw=colOperon]
            coordinates {(58,0.497)};
        \addplot[only marks, mark=triangle*, mark size=5.5pt, fill=colGA, draw=colGA]
            coordinates {(90,0.62)};
        \addplot[only marks, mark=triangle*, mark size=5.5pt, fill=colPSO, draw=colPSO,
                 mark options={rotate=180}]
            coordinates {(90,0.585)};

        \addplot[only marks, mark=o, mark size=5pt, draw=colSLIM, fill=white, thick, forget plot]
            coordinates {(90,0.612)};
        \addplot[only marks, mark=square, mark size=4.5pt, draw=colGPlearn, fill=white, thick, forget plot]
            coordinates {(5,0.538)};
        \addplot[only marks, mark=diamond, mark size=5.5pt, draw=colOperon, fill=white, thick, forget plot]
            coordinates {(42,0.450)};

        \addplot[only marks, mark=star, mark size=5.5pt, draw=colCCI, fill=colCCI]
            coordinates {(90,0.549)};
        \addplot[only marks, mark=+, mark size=5.5pt, ultra thick, draw=colPCCI]
            coordinates {(100,0.560)};

        \end{axis}

        \node[anchor=north, font=\small, inner sep=0pt] at (current axis.below south) {%
            \begin{tabular}{@{}c@{}}
                
                \raisebox{-0.3ex}{\tikz{\color{colSLIM}\pgfsetfillcolor{colSLIM}\pgfsetplotmarksize{5pt}\pgfuseplotmark{*}}}~SLIM
                \hspace{2.5mm}
                \raisebox{-0.3ex}{\tikz{\color{colGPlearn}\pgfsetfillcolor{colGPlearn}\pgfsetplotmarksize{5pt}\pgfuseplotmark{square*}}}~GPlearn
                \hspace{2.5mm}
                \raisebox{-0.3ex}{\tikz{\color{colOperon}\pgfsetfillcolor{colOperon}\pgfsetplotmarksize{5pt}\pgfuseplotmark{diamond*}}}~Operon
                \hspace{2.5mm}
                \raisebox{-0.3ex}{\tikz{\color{black}\pgfsetfillcolor{black}\pgfsetplotmarksize{5pt}\pgfuseplotmark{*}}}~Best perf
                \hspace{5mm}
                \raisebox{-0.3ex}{\tikz{\color{black}\pgfsetplotmarksize{5pt}\pgfuseplotmark{o}}}~Smallest
                \\[3pt]
                \raisebox{-0.3ex}{\tikz{\color{colGA}\pgfsetfillcolor{colGA}\pgfsetplotmarksize{5pt}\pgfuseplotmark{triangle*}}}~GA
                \hspace{2.5mm}
                \raisebox{-0.3ex}{\tikz[rotate=180]{\color{colPSO}\pgfsetfillcolor{colPSO}\pgfsetplotmarksize{5pt}\pgfuseplotmark{triangle*}}}~FST-PSO
                \hspace{2.5mm}
                \raisebox{0.6ex}{\tikz{\draw[colRSF, dash pattern=on 4pt off 3pt, very thick] (0,0)--(0.7,0);}}~RSF
                \hspace{2.5mm}
                \raisebox{0.6ex}{\tikz{\draw[colDSurv, dash pattern=on 4pt off 3pt, very thick] (0,0)--(0.7,0);}}~DeepSurv
                \hspace{2.5mm}
                \raisebox{-0.3ex}{\tikz{\color{colCCI}\pgfsetfillcolor{colCCI}\pgfsetplotmarksize{5pt}\pgfuseplotmark{star}}}~CCI
                \hspace{5mm}
                \raisebox{-0.3ex}{\tikz{\color{colPCCI}\draw[ultra thick] (-3.5pt,0)--(3.5pt,0); \draw[ultra thick] (0,-3.5pt)--(0,3.5pt);}}~PCCI
            \end{tabular}%
        };
    \end{tikzpicture}

    \caption{Two dimensional plot displaying the trade-off between performance (Test C-index) and solution size (nodes) on the standard dataset. The x-axis is inverted for ease of visualization.}
    \label{fig:pareto_standard}
\end{figure*}
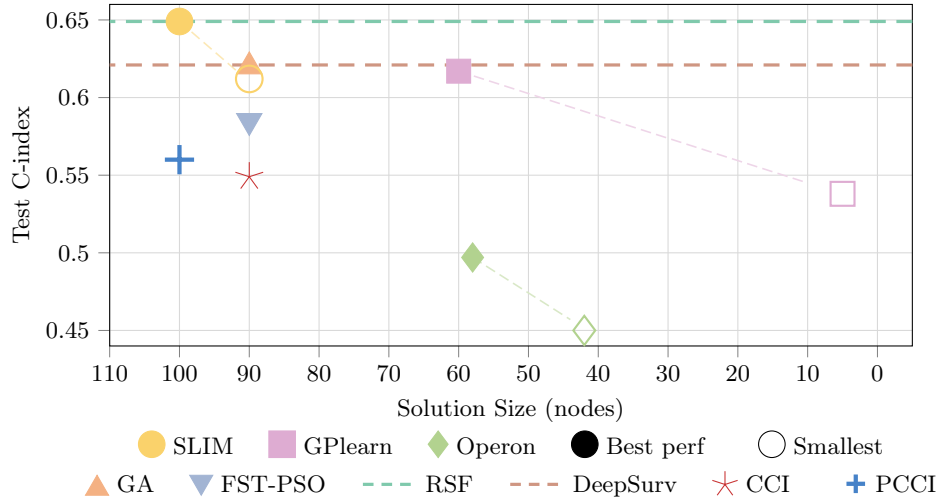

%% file: figures/boxplot_urological.tex
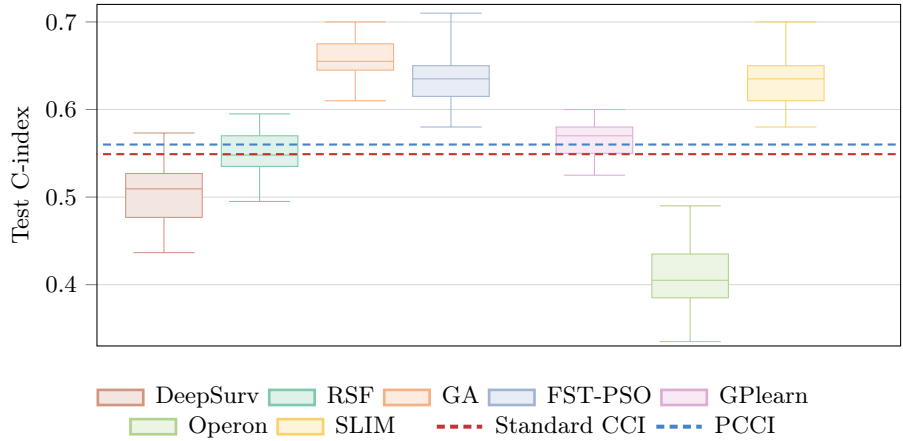
\begin{figure*}[!h]
    \centering
    \begin{tikzpicture}
        \begin{axis}[
            width=\textwidth,
            height=0.5\textwidth,
            gridded,
            noinnerticks,
            ylabel={Test C-index},
            ymin=0.33, ymax=0.72,
            xtick=\empty,
            xmin=-0.7, xmax=7.7,
            boxplot/draw direction=y,
            boxplot/every outlier/.style={mark=*, mark size=1.5pt, mark options={fill=gray!70, draw=gray!70}},
        ]

        \addplot[boxplot prepared={
              lower whisker=0.43655,
              lower quartile=0.47682,
              median=0.50940,
              upper quartile=0.52693,
              upper whisker=0.57319
          }, , fill=colDSurv!25, draw=colDSurv, boxplot/draw position=0] coordinates {};
        
        \addplot[boxplot prepared={
            lower whisker=0.495,
            lower quartile=0.535,
            median=0.548,
            upper quartile=0.570,
            upper whisker=0.595,
        }, fill=colRSF!25, draw=colRSF, boxplot/draw position=1] coordinates {(1,0.47)};

        \addplot[boxplot prepared={
            lower whisker=0.61,
            lower quartile=0.645,
            median=0.655,
            upper quartile=0.675,
            upper whisker=0.70,
        }, fill=colGA!25, draw=colGA, boxplot/draw position=2] coordinates {};

        \addplot[boxplot prepared={
            lower whisker=0.58,
            lower quartile=0.615,
            median=0.635,
            upper quartile=0.650,
            upper whisker=0.71,
        }, fill=colPSO!25, draw=colPSO, boxplot/draw position=3] coordinates {};

        \addplot[boxplot prepared={
            lower whisker=0.525,
            lower quartile=0.55,
            median=0.57,
            upper quartile=0.58,
            upper whisker=0.60,
        }, fill=colGPlearn!25, draw=colGPlearn, boxplot/draw position=4.5] coordinates {(4.5,0.49)(4.5,0.45)};

        \addplot[boxplot prepared={
            lower whisker=0.335,
            lower quartile=0.385,
            median=0.405,
            upper quartile=0.435,
            upper whisker=0.49,
        }, fill=colOperon!25, draw=colOperon, boxplot/draw position=5.5] coordinates {};

        \addplot[boxplot prepared={
            lower whisker=0.58,
            lower quartile=0.61,
            median=0.635,
            upper quartile=0.65,
            upper whisker=0.70,
        }, fill=colSLIM!25, draw=colSLIM, boxplot/draw position=6.5] coordinates {};

        \addplot[colCCI, densely dashed, thick] coordinates {(-3,0.549) (7.7,0.549)};
        \addplot[colPCCI, densely dashed, thick] coordinates {(-3,0.560) (7.7,0.560)};

        \end{axis}
    \end{tikzpicture}

    \addlegendimageintext{draw=colDSurv, fill=colDSurv!25, area legend, line width=1pt} DeepSurv
    \addlegendimageintext{draw=colRSF, fill=colRSF!25, area legend, line width=1pt} RSF
    \addlegendimageintext{draw=colGA, fill=colGA!25, area legend, line width=1pt} GA
    \addlegendimageintext{draw=colPSO, fill=colPSO!25, area legend, line width=1pt} FST-PSO 
    \addlegendimageintext{draw=colGPlearn, fill=colGPlearn!25, area legend, line width=1pt} GPlearn \\
    \addlegendimageintext{draw=colOperon, fill=colOperon!25, area legend, line width=1pt} Operon
    \addlegendimageintext{draw=colSLIM, fill=colSLIM!25, area legend, line width=1pt} SLIM
    \hspace{3mm} 
    \addlegendimageintext{draw=colCCI, densely dashed, line width=1.5pt} Standard CCI
    \addlegendimageintext{draw=colPCCI, densely dashed, line width=1.5pt} PCCI

    \caption{Test C-index comparison across all methods when including PCa specific variables in the dataset.}
    \label{fig:boxplot_urological}
\end{figure*}

%% file: figures/boxplot_size_urological.tex
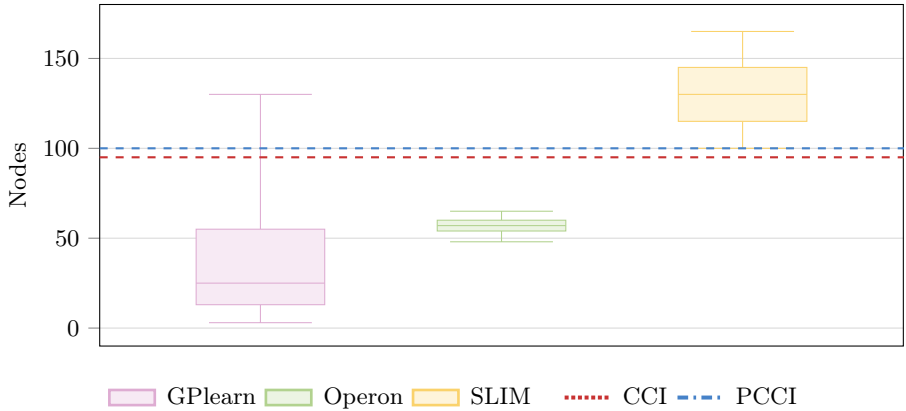
\begin{figure*}[!h]
    \centering
    \begin{tikzpicture}
        \begin{axis}[
            width=\textwidth,
            height=0.5\textwidth,
            gridded,
            noinnerticks,
            ylabel={Nodes},
            ymin=-10, ymax=180,
            xtick=\empty,
            xmin=0, xmax=5,
            boxplot/draw direction=y,
            boxplot/every outlier/.style={mark=*, mark size=1.5pt, mark options={fill=gray!70, draw=gray!70}},
        ]

        \addplot[boxplot prepared={
            lower whisker=3,
            lower quartile=13,
            median=25,
            upper quartile=55,
            upper whisker=130,
        }, fill=colGPlearn!25, draw=colGPlearn, boxplot/draw position=1] coordinates {};

        \addplot[boxplot prepared={
            lower whisker=48,
            lower quartile=54,
            median=57,
            upper quartile=60,
            upper whisker=65,
        }, fill=colOperon!25, draw=colOperon, boxplot/draw position=2.5] coordinates {};

        \addplot[boxplot prepared={
            lower whisker=100,
            lower quartile=115,
            median=130,
            upper quartile=145,
            upper whisker=165,
        }, fill=colSLIM!25, draw=colSLIM, boxplot/draw position=4] coordinates {};

        \addplot[colCCI, dashed,  thick] coordinates {(0,95) (6.5,95)};
        \addplot[colPCCI, dashed, thick] coordinates {(0,100) (6.5,100)};

        \end{axis}
    \end{tikzpicture}

    \addlegendimageintext{draw=colGPlearn, fill=colGPlearn!25, area legend, line width=1pt} GPlearn
    \addlegendimageintext{draw=colOperon, fill=colOperon!25, area legend, line width=1pt} Operon
    \addlegendimageintext{draw=colSLIM, fill=colSLIM!25, area legend, line width=1pt} SLIM
    \hspace{2mm}
    \addlegendimageintext{draw=colCCI, densely dotted, line width=1.5pt} CCI
    \addlegendimageintext{draw=colPCCI, dash pattern=on 4pt off 2pt on 1pt off 2pt, line width=1.5pt} PCCI

    \caption{Model size comparison (number of nodes) on the urological variables dataset.}
    \label{fig:boxplot_size_urological}
\end{figure*}

%% file: figures/pareto_urological.tex
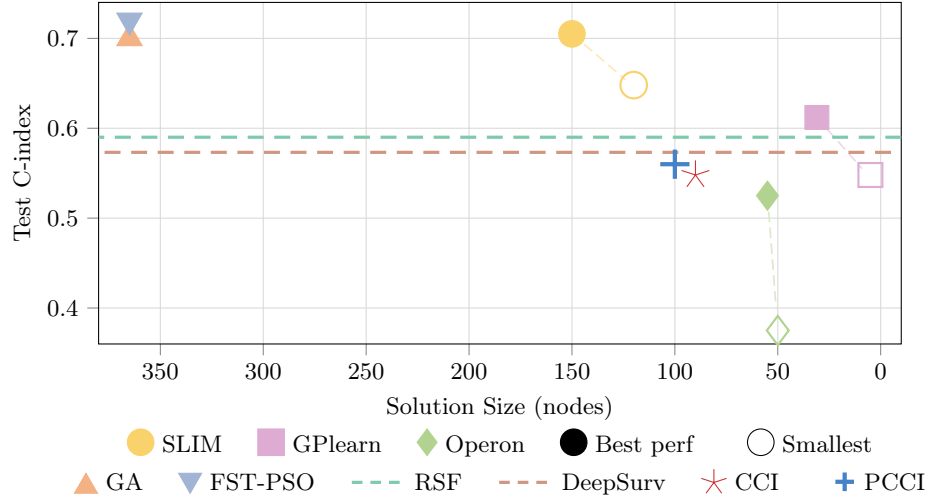
\begin{figure*}[!h]
    \centering
    \begin{tikzpicture}
        \begin{axis}[
            width=\textwidth,
            height=0.5\textwidth,
            gridded,
            noinnerticks,
            xlabel={Solution Size (nodes)},
            ylabel={Test C-index},
            ymin=0.36, ymax=0.74,
            xmin=-10, xmax=380,
            x dir=reverse,
            legend style={hide legend},
        ]

        \addplot[colSLIM!50, dash pattern=on 4pt off 2pt, semithick, forget plot]
            coordinates {(150,0.705) (124,0.654)};
        \addplot[colGPlearn!50, dash pattern=on 4pt off 2pt, semithick, forget plot]
            coordinates {(31,0.612) (9,0.555)};
        \addplot[colOperon!50, dash pattern=on 4pt off 2pt, semithick, forget plot]
            coordinates {(55,0.525) (50.8,0.400)};

        \addplot[colRSF, dash pattern=on 6pt off 4pt, very thick]
            coordinates {(-10,0.59) (380,0.59)};
        \addplot[colDSurv, dash pattern=on 6pt off 4pt, very thick]
            coordinates {(-5,0.57319) (380,0.57319)};

        \addplot[only marks, mark=*, mark size=5pt, fill=colSLIM, draw=colSLIM]
            coordinates {(150,0.705)};
        \addplot[only marks, mark=square*, mark size=4.5pt, fill=colGPlearn, draw=colGPlearn]
            coordinates {(31,0.612)};
        \addplot[only marks, mark=diamond*, mark size=5.5pt, fill=colOperon, draw=colOperon]
            coordinates {(55,0.525)};
        \addplot[only marks, mark=triangle*, mark size=5.5pt, fill=colGA, draw=colGA]
            coordinates {(365,0.7)};
        \addplot[only marks, mark=triangle*, mark size=5.5pt, fill=colPSO, draw=colPSO,
                 mark options={rotate=180}]
            coordinates {(365,0.72)};

        \addplot[only marks, mark=o, mark size=5pt, draw=colSLIM, fill=white, thick, forget plot]
            coordinates {(120,0.648)};
        \addplot[only marks, mark=square, mark size=4.5pt, draw=colGPlearn, fill=white, thick, forget plot]
            coordinates {(5,0.548)};
        \addplot[only marks, mark=diamond, mark size=5.5pt, draw=colOperon, fill=white, thick, forget plot]
            coordinates {(50,0.375)};

        \addplot[only marks, mark=star, mark size=5.5pt, draw=colCCI, fill=colCCI]
            coordinates {(90,0.548)};
        \addplot[only marks, mark=+, mark size=5.5pt, ultra thick, draw=colPCCI]
            coordinates {(100,0.560)};

        \end{axis}

        \node[anchor=north, font=\small, inner sep=0pt] at (current axis.below south) {%
            \begin{tabular}{@{}c@{}}
                
                \raisebox{-0.3ex}{\tikz{\color{colSLIM}\pgfsetfillcolor{colSLIM}\pgfsetplotmarksize{5pt}\pgfuseplotmark{*}}}~SLIM
                \hspace{2.5mm}
                \raisebox{-0.3ex}{\tikz{\color{colGPlearn}\pgfsetfillcolor{colGPlearn}\pgfsetplotmarksize{5pt}\pgfuseplotmark{square*}}}~GPlearn
                \hspace{2.5mm}
                \raisebox{-0.3ex}{\tikz{\color{colOperon}\pgfsetfillcolor{colOperon}\pgfsetplotmarksize{5pt}\pgfuseplotmark{diamond*}}}~Operon
                \hspace{2.5mm}
                \raisebox{-0.3ex}{\tikz{\color{black}\pgfsetfillcolor{black}\pgfsetplotmarksize{5pt}\pgfuseplotmark{*}}}~Best perf
                \hspace{5mm}
                \raisebox{-0.3ex}{\tikz{\color{black}\pgfsetplotmarksize{5pt}\pgfuseplotmark{o}}}~Smallest
                \\[3pt]
                \raisebox{-0.3ex}{\tikz{\color{colGA}\pgfsetfillcolor{colGA}\pgfsetplotmarksize{5pt}\pgfuseplotmark{triangle*}}}~GA
                \hspace{2.5mm}
                \raisebox{-0.3ex}{\tikz[rotate=180]{\color{colPSO}\pgfsetfillcolor{colPSO}\pgfsetplotmarksize{5pt}\pgfuseplotmark{triangle*}}}~FST-PSO
                \hspace{2.5mm}
                \raisebox{0.6ex}{\tikz{\draw[colRSF, dash pattern=on 4pt off 3pt, very thick] (0,0)--(0.7,0);}}~RSF
                \hspace{2.5mm}
                \raisebox{0.6ex}{\tikz{\draw[colDSurv, dash pattern=on 4pt off 3pt, very thick] (0,0)--(0.7,0);}}~DeepSurv
                \hspace{2.5mm}
                \raisebox{-0.3ex}{\tikz{\color{colCCI}\pgfsetfillcolor{colCCI}\pgfsetplotmarksize{5pt}\pgfuseplotmark{star}}}~CCI
                \hspace{5mm}
                \raisebox{-0.3ex}{\tikz{\color{colPCCI}\draw[ultra thick] (-3.5pt,0)--(3.5pt,0); \draw[ultra thick] (0,-3.5pt)--(0,3.5pt);}}~PCCI
            \end{tabular}%
        };
    \end{tikzpicture}

    \caption{Two dimensional plot displaying the trade-off between performance (Test C-index) and solution size (nodes) on the standard dataset. The x-axis is inverted for ease of visualization.}
    \label{fig:pareto_urological}
\end{figure*}

%% file: bibl.bib
@article{cci, title={A new method of classifying prognostic comorbidity in longitudinal studies: Development and validation}, volume={40}, url={https://pubmed.ncbi.nlm.nih.gov/3558716/}, DOI={https://doi.org/10.1016/0021-9681(87)90171-8}, number={5}, journal={Journal of Chronic Diseases}, publisher={Elsevier BV}, author={Charlson, Mary E. and Pompei, Peter and Ales, Kathy L. and MacKenzie, C.Ronald}, year={1987}, month={Jan}, pages={373–383} }

@article{pcci, title={An Age Adjusted Comorbidity Index to Predict Long-Term, Other Cause Mortality in Men with Prostate Cancer}, volume={194}, url={https://pubmed.ncbi.nlm.nih.gov/25623745/}, DOI={https://doi.org/10.1016/j.juro.2015.01.081}, number={1}, journal={Journal of Urology}, publisher={Ovid Technologies (Wolters Kluwer Health)}, author={Daskivich, Timothy J. and Kwan, Lorna and Dash, Atreya and Saigal, Christopher and Litwin, Mark S.}, year={2015}, month={Jul}, pages={73–78} }

@article{acci, title={Development of an algorithm for the Charlson Comorbidity Index using the National Database Study in the Japan Kanto 7 Prefectures}, url={https://www.medrxiv.org/content/10.1101/2025.04.10.25325568v1}, DOI={https://doi.org/10.1101/2025.04.10.25325568}, publisher={openRxiv}, author={Sekine, Airi and Nakajima, Kei}, year={2025}, month={Apr} }

@article{Siegel2023CancerStats, title={Cancer statistics, 2023}, volume={73}, url={https://pubmed.ncbi.nlm.nih.gov/36633525/}, DOI={https://doi.org/10.3322/caac.21763}, number={1}, journal={CA: A Cancer Journal for Clinicians}, publisher={Wiley}, author={Siegel, Rebecca L. and Miller, Kimberly D. and Wagle, Nikita Sandeep and Jemal, Ahmedin}, year={2023}, month={Jan}, pages={17–48} }

@article{Hamdy2016ProtecT, title={10-Year Outcomes after Monitoring, Surgery, or Radiotherapy for Localized Prostate Cancer}, volume={375}, url={https://pubmed.ncbi.nlm.nih.gov/27626136/}, DOI={https://doi.org/10.1056/nejmoa1606220}, number={15}, journal={New England Journal of Medicine}, publisher={Massachusetts Medical Society}, author={Hamdy, Freddie C. and Donovan, Jenny L. and Lane, J. Athene and Mason, Malcolm and Metcalfe, Chris and Holding, Peter and Davis, Michael and Peters, Tim J. and Turner, Emma L. and Martin, Richard M. and Oxley, Jon and Robinson, Mary and Staffurth, John and Walsh, Eleanor and Bollina, Prasad and Catto, James and Doble, Andrew and Doherty, Alan and Gillatt, David and Kockelbergh, Roger}, year={2016}, month={Oct}, pages={1415–1424} }

@article{EAU2024Guidelines,
  title={{EAU-EANM-ESTRO-ESUR-ISUP-SIOG} Guidelines on Prostate Cancer—2024 Update. Part I: Screening, Diagnosis, and Local Treatment with Curative Intent},
  volume={86},
  url={https://pubmed.ncbi.nlm.nih.gov/38614820/},
  DOI={https://doi.org/10.1016/j.eururo.2024.03.027},
  number={2},
  journal={European Urology},
  publisher={Elsevier BV},
  author={Cornford, Philip and van den Bergh, Roderick C.N. and Briers, Erik and Van den Broeck, Thomas and Brunckhorst, Oliver and Darraugh, Julie and Eberli, Daniel and De Meerleer, Gert and De Santis, Maria and Farolfi, Andrea and Gandaglia, Giorgio and Gillessen, Silke and Grivas, Nikolaos and Henry, Ann M. and Lardas, Michael and van Leenders, Geert J.L.H. and Liew, Matthew and Linares Espinos, Estefania and Oldenburg, Jan and van Oort, Inge M. and Ploussard, Guillaume and Roberts, Matthew J. and Rouvière, Olivier and Schoots, Ivo G. and Schouten, Natasha and Smith, Emma J. and Stranne, Johan and Wiegel, Thomas and Willemse, Peter-Paul M. and Tilki, Derya},
  year={2024},
  month={Aug},
  pages={148–163}
}

@article{Palella1998HAART, title={Declining Morbidity and Mortality among Patients with Advanced Human Immunodeficiency Virus Infection}, volume={338}, url={https://pubmed.ncbi.nlm.nih.gov/9516219/}, DOI={https://doi.org/10.1056/nejm199803263381301}, number={13}, journal={New England Journal of Medicine}, publisher={Massachusetts Medical Society}, author={Palella, Frank J. and Delaney, Kathleen M. and Moorman, Anne C. and Loveless, Mark O. and Fuhrer, Jack and Satten, Glen A. and Aschman, Diane J. and Holmberg, Scott D.}, year={1998}, month={Mar}, pages={853–860} }

@article{Samji2013HIVLifeExpectancy, title={Closing the Gap: Increases in Life Expectancy among Treated HIV-Positive Individuals in the United States and Canada}, volume={8}, url={https://pubmed.ncbi.nlm.nih.gov/24367482/}, DOI={https://doi.org/10.1371/journal.pone.0081355}, number={12}, journal={PLoS ONE}, publisher={Public Library of Science (PLoS)}, author={Samji, Hasina and Cescon, Angela and Hogg, Robert S. and Modur, Sharada P. and Althoff, Keri N. and Buchacz, Kate and Burchell, Ann N. and Cohen, Mardge and Gebo, Kelly A. and Gill, M. John and Justice, Amy and Kirk, Gregory and Klein, Marina B. and Korthuis, P. Todd and Martin, Jeff and Napravnik, Sonia and Rourke, Sean B. and Sterling, Timothy R. and Silverberg, Michael J. and Deeks, Stephen}, editor={Okulicz, Jason F.}, year={2013}, month={Dec}, pages={e81355} }

@Book{GA,
author = "Leonardo Vanneschi and Sara Silva",
title = "Lectures on Intelligent Systems",
publisher = "Springer",
year = "2023",
series = "Genetic and Evolutionary Computation",
isbn13 = "978-3-031-17921-1",
ISSN = "1932-0175",
ISSN = "1619-7127",
chapter = "Genetic Algorithm"}

@article{fst_pso, title={Fuzzy Self-Tuning PSO: A settings-free algorithm for global optimization}, volume={39}, url={https://www.sciencedirect.com/science/article/pii/S2210650216303534}, DOI={https://doi.org/10.1016/j.swevo.2017.09.001}, journal={Swarm and Evolutionary Computation}, publisher={Elsevier BV}, author={Nobile, Marco S. and Cazzaniga, Paolo and Besozzi, Daniela and Colombo, Riccardo and Mauri, Giancarlo and Pasi, Gabriella}, year={2018}, month={Apr}, pages={70–85} }

@inproceedings{operon,
    author = {Burlacu, Bogdan and Kronberger, Gabriel and Kommenda, Michael},
    title = {Operon C++: An Efficient Genetic Programming Framework for Symbolic Regression},
    year = {2020},
    isbn = {9781450371278},
    publisher = {Association for Computing Machinery},
    address = {New York, NY, USA},
    url = {https://doi.org/10.1145/3377929.3398099},
    doi = {10.1145/3377929.3398099},
    booktitle = {Proceedings of the 2020 Genetic and Evolutionary Computation Conference Companion},
    pages = {1562–1570},
    numpages = {9},
    location = {Canc\'{u}n, Mexico},
    series = {GECCO '20}
}

@misc{lopez_2026, title={gplearn-readthedocs-io-en-stable}, url={https://www.scribd.com/document/691097226/gplearn-readthedocs-io-en-stable}, journal={Scribd}, author={lopez, juan}, year={2026} }

@article{slim, address={Singapore}, title={Exploring Non-bloating Geometric Semantic Genetic Programming}, ISBN={9789819600762}, url={https://link.springer.com/chapter/10.1007/978-981-96-0077-9_12}, DOI={https://doi.org/10.1007/978-981-96-0077-9_12}, journal={Genetic and Evolutionary Computation}, publisher={Springer Nature Singapore}, author={Vanneschi, Leonardo and Farinati, Davide and Rasteiro, Diogo and Rosenfeld, Liah and Pietropolli, Gloria and Silva, Sara}, year={2025}, pages={237–258} }

@article{slim_impl, address={New York, NY, USA}, title={Slim\_gsgp: A Python Library for Non-Bloating GSGP}, DOI={https://doi.org/10.1145/3712256.3726398}, journal={Proceedings of the Genetic and Evolutionary Computation Conference}, publisher={ACM}, author={Rosenfeld, Liah and Farinati, Davide and Rasteiro, Diogo and Pietropolli, Gloria and Rebuli, Karina and Silva, Sara and Vanneschi, Leonardo}, year={2025}, month={Jul}, pages={1026–1034} }

@article{rsf, title={Random survival forests}, volume={2}, url={https://projecteuclid.org/journals/annals-of-applied-statistics/volume-2/issue-3/Random-survival-forests/10.1214/08-AOAS169.full}, DOI={https://doi.org/10.1214/08-aoas169}, number={3}, journal={The Annals of Applied Statistics}, publisher={Institute of Mathematical Statistics}, author={Ishwaran, Hemant and Kogalur, Udaya B. and Blackstone, Eugene H. and Lauer, Michael S.}, year={2008}, month={Sep} }

@article{farinati_vanneschi_2024, title={A survey on dynamic populations in bio-inspired algorithms}, volume={25}, url={https://link.springer.com/article/10.1007/s10710-024-09492-4}, DOI={https://doi.org/10.1007/s10710-024-09492-4}, number={2}, journal={Genetic Programming and Evolvable Machines}, publisher={Springer Science and Business Media LLC}, author={Farinati, Davide and Vanneschi, Leonardo}, year={2024}, month={Jul} }

@article{charlson1994, title={Validation of a combined comorbidity index}, volume={47}, url={https://pubmed.ncbi.nlm.nih.gov/7722560/}, DOI={https://doi.org/10.1016/0895-4356(94)90129-5}, number={11}, journal={Journal of Clinical Epidemiology}, publisher={Elsevier BV}, author={Charlson, Mary and Szatrowski, Ted P. and Peterson, Janey and Gold, Jeffrey}, year={1994}, month={Nov}, pages={1245–1251} }

@article{farinati_pietropolli_vanneschi_2025, title={A Study on the Dynamics and Effectiveness of the Deflate Geometric Semantic Mutation}, url={https://ieeexplore.ieee.org/abstract/document/11168418}, DOI={https://doi.org/10.1109/tevc.2025.3611226}, journal={IEEE Transactions on Evolutionary Computation}, publisher={Institute of Electrical and Electronics Engineers (IEEE)}, author={Farinati, Davide and Pietropolli, Gloria and Vanneschi, Leonardo}, year={2025}, pages={1–1} }

@article{deepsurv, title={DeepSurv: personalized treatment recommender system using a Cox proportional hazards deep neural network}, volume={18}, url={https://link.springer.com/article/10.1186/s12874-018-0482-1}, DOI={https://doi.org/10.1186/s12874-018-0482-1}, number={1}, journal={BMC Medical Research Methodology}, publisher={Springer Science and Business Media LLC}, author={Katzman, Jared L. and Shaham, Uri and Cloninger, Alexander and Bates, Jonathan and Jiang, Tingting and Kluger, Yuval}, year={2018}, month={Feb} }

@article{cox_oakes_2018, title={Analysis of Survival Data}, ISBN={9781315137438}, url={https://www.taylorfrancis.com/books/mono/10.1201/9781315137438/analysis-survival-data-reid-cox}, DOI={https://doi.org/10.1201/9781315137438}, publisher={Chapman and Hall/CRC}, author={Cox, D.R. and Oakes, D.}, year={2018}, month={Feb} }
